\documentclass[runningheads]{llncs}

\usepackage[T1]{fontenc}
\usepackage{pbox}
\usepackage{graphicx}
\usepackage{subcaption}
\usepackage{url}
\usepackage{amsmath}

\begin{document}

\title{EquiSumm : A Gender Bias-Aware Framework for Inclusive Tweet Summarization}
\titlerunning{EquiSumm}

\author{Chaitanya Wanjari\inst{1} \and
Jessica Kamal\inst{1} \and
Riddhi Jain\inst{1}\and
Samruddhi Kurhe\inst{1} \and
Roshni Chakraborty\inst{1}}

\authorrunning{EquiSumm}

\institute{ABV IIITM Gwalior, India\\
\email{\{imt\_2022034, imt\_2022054, imt\_2022097, imt\_2022100, roshni\}@iiitm.ac.in}}

\maketitle

\begin{abstract}
While social media platforms, such as Twitter, provide a medium for large-scale opinion sharing during news events, it is manually impossible for individuals or media agencies to process the vast volume of content to identify key viewpoints. In order to resolve this, several automatic summarization techniques have been proposed to condense large collections of tweets into concise and informative summaries. However, these algorithms do not explicitly consider demographic fairness. Several existing research works have developed automated summarization approaches that can provide a holistic overview of the key aspects and major opinions shared on social media platforms related to a news event. However, these approaches do not explicitly consider different forms of demographic representation, such as gender, which can lead to biased summary representation. In this paper, we propose \textit{EquiSumm}, which considers the gender aspect of the shared opinion to generate a summary, and our experimental analysis on two major datasets indicates the performance effectiveness with respect to existing research works.
\keywords{gender aware \and Tweet Summarization \and fairness}
\end{abstract}

\section{Introduction}
Although social media platforms, such as Twitter, have become the main sources of information for a large fraction of users, the availability of continuous information is overwhelming to identify and understand the aspects. Therefore, automated summarization approaches can provide an overview of user opinions with respect to a news event through a small number of tweets~\cite{chakraborty2019tweet}. A comprehensive summary of user opinions can provide users an understanding of the larger debate centered and provide representation to the different viewpoints ~\cite{garg2025portrait}, such as position detection ~\cite{chakraborty2022detecting}, popularity identification, and holistic understanding of the event \cite{zhu2025eventsum,mahindrakar2024performance,kumar2024extracting}. There are several challenges in tweet summarization, such as understanding of user viewpoints irrespective of the vocabulary gap, etc \cite{bansal2021dcbrts}. While news event based tweet summarization focuses on understanding the \textit{relevance} and \textit{coverage} of summary tweets with respect to the news event \cite{chakraborty2023twminer}, disaster tweet based summarization approaches mainly aim to identify the different subcategories to ensure faster disaster response \cite{garg2025atsumm}. However, none of these approaches focus on social bias aware summarization, i.e. integration and representation of the different groups of individuals.

\par Therefore, while existing summarization systems are widely used across news media platforms, social media applications provide users with an overview of what content is highlighted and how it is presented. However, these approaches might not ensure the representation of all demographic groups. This becomes specifically relevant when news events relate directly to demographic groups, such as, \textit{MeToo} event\footnote{\url{https://en.wikipedia.org/wiki/MeToo_movement}}. Therefore, an appropriate summary should ensure both representation of the opinions and fair representation of both genders.

\begin{table}
\caption{Tweets with respect to \textit{MeToo Dataset}}
\label{tab:uR2}
\centering
\begin{tabular}{|c|p{10cm}|c|}
\hline
\textbf{SNo} & \textbf{Tweet} & \textbf{$GC$} \\
\hline
1 & \pbox{12cm}{\textit{In the age of \#MeToo and biased laws, men are suffering...}} & $M$\\
\hline
2 & \pbox{12cm}{\textit{Has any woman from biological science background cried victim...?}} & $F$\\
\hline
3 & \pbox{12cm}{\textit{Pain knows no gender. When it hurts, it hurts equally...}} & $B$\\
\hline
4 & \pbox{12cm}{\textit{With the rise of \#MeToo, workplace dynamics are changing...}} & $N$\\
\hline
\end{tabular}
\end{table}

\section{Proposed Methodology} The proposed framework comprises of two phases, which includes classification of a tweet to a particular gender group followed by selection of representative tweets from each group. We discuss each of these steps in detail next. 
\subsubsection*{Phase I (Gender Classification) :} This is the first step in the summarization method. The main objective for this Phase is to identify the aspects related to the event that the specific gender discusses and further, would ensure representation for every gender in summarization. Our Phase I comprises of classification of a tweet into one of the gender categories ($GC$), such as, \textit{male} ($M$) , \textit{female} ($F$), \textit{neutral} ($N$), or \textit{both} ($B$), on the basis of the topic and information discussed in that tweet text. We provide four examples to highlight the understanding of the different aspects to identify the gender that the tweet implicitly discusses in Table \ref{tab:uR2}. The proposed approach does not consider the gender of the user who tweeted rather proposes an approach to automatically infer the particular gender the tweet refers to or discusses about. Therefore, understanding and categorizing tweets to different genders on the basis of their \textit{implicit} references would help to ensure opinions and information representativeness across different genders. \par
\begin{figure}[h]
\centering

\begin{subfigure}[b]{0.48\linewidth}
\centering
\includegraphics[width=\linewidth]{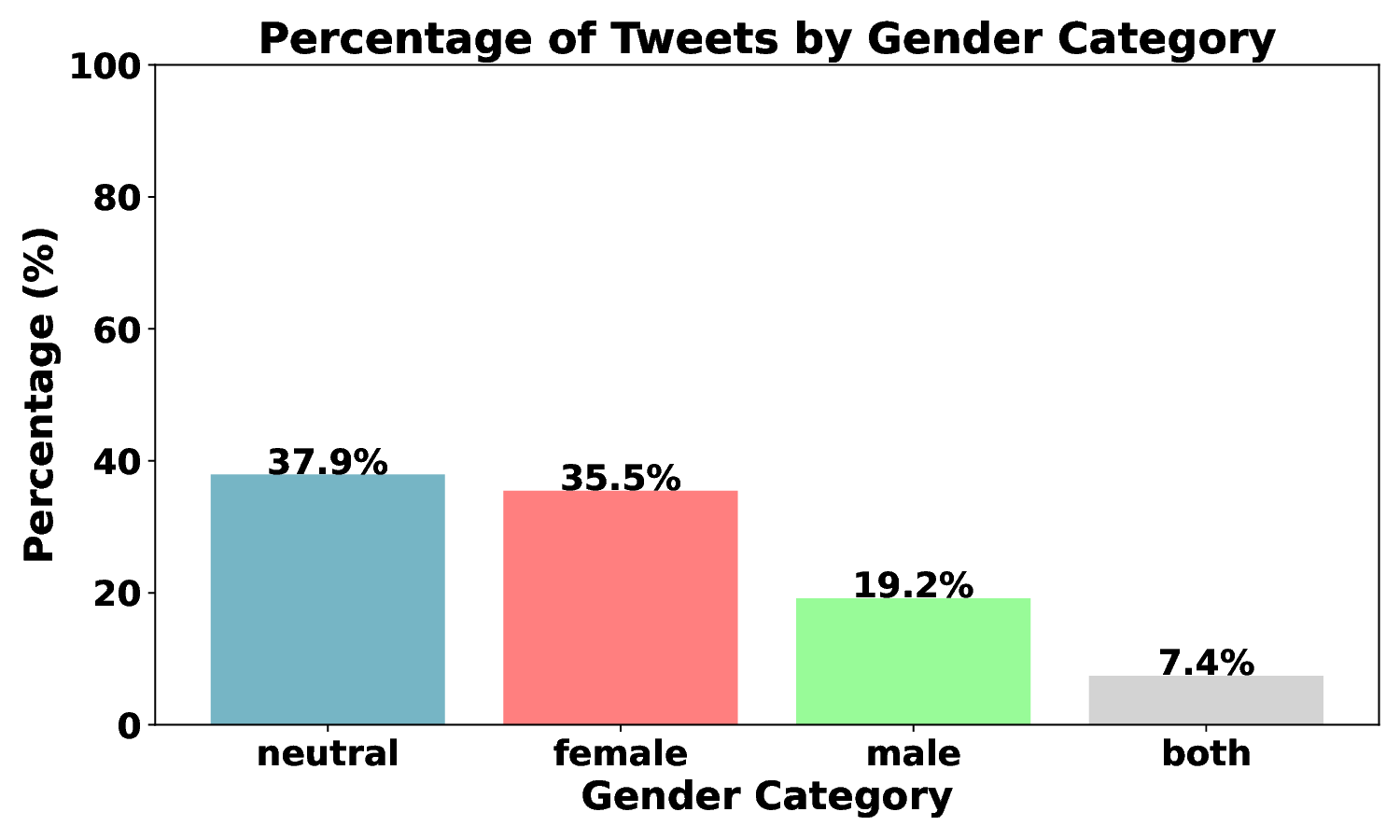}
\caption{MeToo Dataset}
\label{fig:g1}
\end{subfigure}
\hfill
\begin{subfigure}[b]{0.48\linewidth}
\centering
\includegraphics[width=\linewidth]{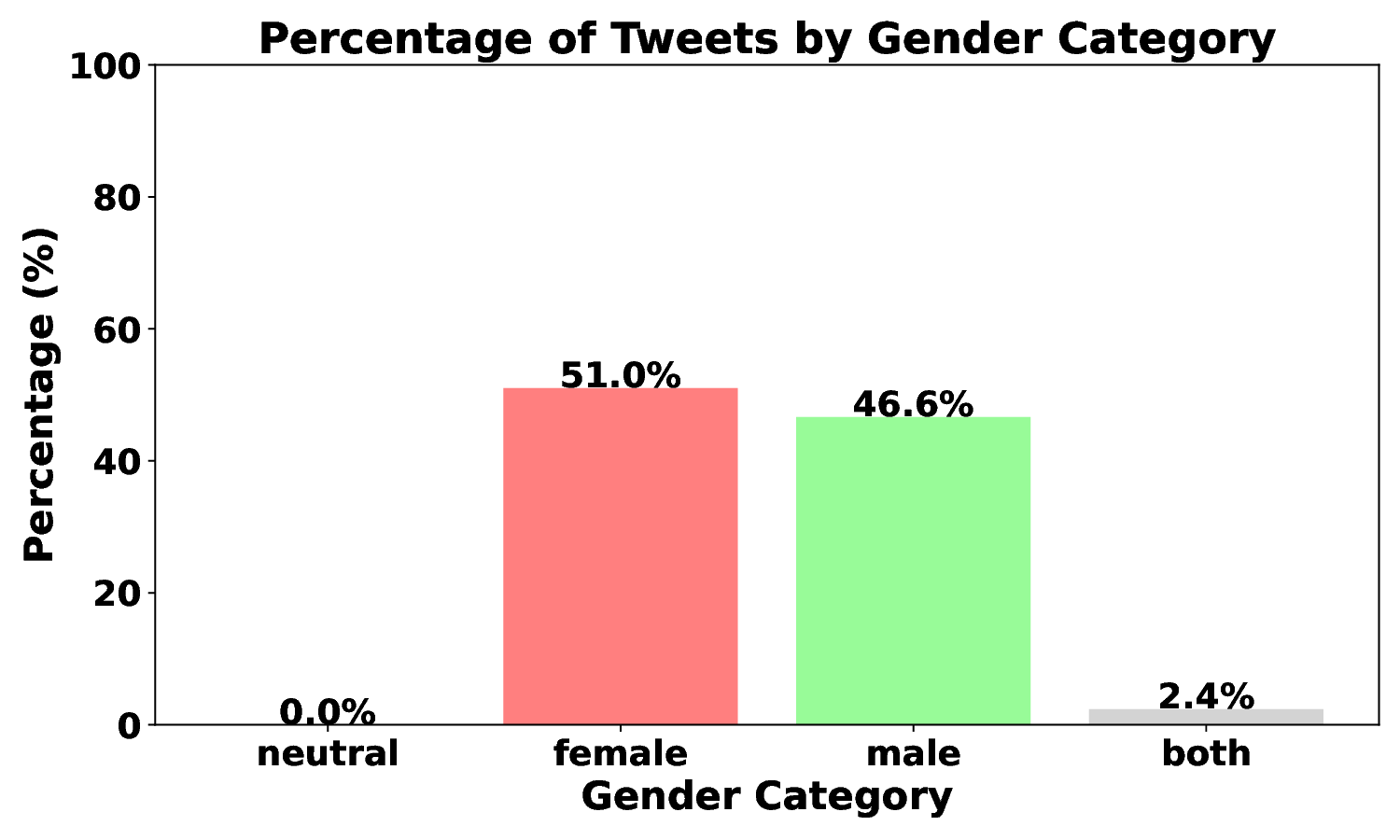}
\caption{Legalisation of Abortion Dataset}
\label{fig:g2}
\end{subfigure}

\caption{Percentage of tweets across gender categories obtained using spaCy NER and clustering.}
\end{figure}

 For Phase I, we propose a clustering based technique to identify the gender category given a tweet. We initially classify a tweet to a particular gender on the basis of its constituent word matching with the existing gender ontology. We rely on the word list provided by the MIND dataset with respect to male and female-associated words \cite{gb2025}. We follow spaCy-NER\footnote{https://spacy.io/} to detect gendered mentions in tweets such as names typically associated with men or women or gendered pronouns. Incorporating gender-specific keywords drawn from the existing ontology with spaCy’s NER, we get tweets classified into gender categories. This further includes understanding of gender specific names along with contextual information. If the tweet contains words from both the male and female lists, it is considered \textit{Both}. If the number of words from the male list and female list is equal or zero, then the tweet is categorized as \textit{Neutral}. On the basis of our initial segregation, we generate the classified tweets with high confidence score into gender based clusters and compute the corresponding \textit{centroid}, which is the average vector of all tweets in that group on the basis of SBERT \cite{reimers2019sentence} similarity as shown in Equation \ref{eq:gC} \begin{eqnarray} \vec{c}_{\text{gender}} = \frac{1}{N_g} \sum_{i=1}^{N_g} \vec{x}_i \label{eq:gC} \end{eqnarray} \\ Where, $\vec{x}_i$ is the SBERT vector of the tweet $i$ labeled any of the gender, and $N_g$ is the number of tweets in that gender group. Furthermore, given an unclassified tweet or classified tweet with low confidence, we compare the vector representations of that tweet to the corresponding gender cluster centroids by \textit{cosine similarity} \footnote{\url{https://en.wikipedia.org/wiki/Cosine_similarity}}
. An unclassified tweet is then assigned to the gender group to which it is closest in terms of cosine similarity. Therefore, tweets are classified into gender categories like \textbf{male}, \textbf{female}, and \textbf{both}. Our initial results show high effectiveness in gender classification as shown in Figures \ref{fig:g1} and \ref{fig:g2}, respectively, for both the datasets. 

\subsubsection*{Phase II (Representative Tweets Selection)}
In the second phase, the goal is to identify the most informative and representative tweets from each gender-associated cluster so that the final summary reflects both the key discussion themes and balanced gender perspectives. To achieve this, we first construct a tweet similarity graph for each gender group. In this graph, each tweet is represented as a node, and an edge is introduced between two nodes if the cosine similarity between their SBERT-based vector embeddings exceeds a manually chosen threshold. Based on feedback from human annotators and empirical inspection of semantic coherence, we set this threshold to 0.40. This ensures that edges reflect meaningful semantic similarity rather than superficial lexical overlap.

Once the similarity graph is constructed, we apply the LexRank centrality algorithm to identify the most central tweets within each gender cluster. LexRank ranks tweets based on their connectivity and importance within the similarity network, allowing us to select the tweets that best represent the overall viewpoint distribution within that gender group. For each gender category, we extract the top $K$ highest-ranked tweets, where $K$ is fixed to maintain consistency in the length of the generated summaries across gender groups. This prevents the dominant gender group in the dataset from disproportionately influencing the final summary.

Finally, the selected representative tweets from each gender category are concatenated to form the overall event summary. By ensuring equal contribution from each gender-associated cluster, the resulting summary not only captures the key themes and sentiments discussed in the event but also ensures balanced representation of gender perspectives.

\section{Experimental Discussions and Results}

\textbf{Comparison with Baselines.}
We compare \textit{EquiSumm} against widely-used extractive summarization baselines to assess both summary quality and fairness. The first baseline, \textbf{LexRank} \cite{erkan2004lexrank}, is a graph-based centrality method that selects sentences based on their importance within a similarity graph. The second baseline, \textbf{Latent Semantic Analysis (LSA)} \cite{landauer1997solution}, performs dimensionality reduction to uncover latent conceptual topics and selects representative sentences accordingly. The third baseline, \textbf{Community Detection (Louvain) + LexRank}, first clusters tweets into semantically coherent communities and then applies LexRank within each cluster to extract representative sentences. This method allows topic-wise coverage but does not explicitly ensure demographic representation. These baselines help us evaluate how conventional summarizers behave in the presence of gender-skewed discourse.

\textbf{Dataset Details.}
We conduct experiments on two publicly available social discussion datasets where gender-based perspectives are strongly articulated. The first dataset concerns the global \textbf{\#MeToo Movement} and contains 485 tweets reflecting personal experiences, opinions, and reactions to harassment-related narratives. The second dataset captures the debate over the \textbf{Legalization of Abortion in the United States}, consisting of 934 tweets discussing rights, ethics, policy decisions, and personal viewpoints. Both datasets naturally contain diverse emotional tones, argument styles, and implicit references to gender, making them suitable for evaluating fairness in summarization.

\textbf{Inclusion Bias Score.}
Since traditional summarization evaluation metrics such as ROUGE require reference summaries and do not account for demographic fairness, we adopt the \textbf{Inclusion Bias Score (IBS)} to measure the extent of gender representation balance in the generated summaries. IBS measures how frequently male-associated terms appear relative to female-associated terms in the final summary. A score close to 0 indicates balanced representation, while a positive score reflects bias toward female-associated content and a negative score indicates bias toward male-associated content. The metric is computed as follows:

\[
\text{IBS} = 
\frac{\sum freq(f)}{\sum freq(m) + \sum freq(f)} -
\frac{\sum freq(m)}{\sum freq(m) + \sum freq(f)}
\]

where $\text{freq}(m)$ and $\text{freq}(f)$ denote the normalized frequencies of male-associated and female-associated terms, respectively, in the generated summary. This provides a quantitative measure to examine whether the summarization method amplifies or reduces existing bias within the dataset.

\begin{table}[h!]
\centering
\small
\begin{tabular}{|l|c|c|c|c|}
\hline
Approach & G1 & G2 & Score & Direction \\
\hline
Dataset & 0.340 & 0.650 & +0.310 & G2 \\
LexRank & 0.300 & 0.690 & +0.390 & G2 \\
LSA & 0.397 & 0.603 & +0.206 & G2 \\
Community+LexRank & 0.320 & 0.670 & +0.350 & G2 \\
\textit{EquiSumm} & 0.520 & 0.470 & –0.050 & G1 \\
\hline
\end{tabular}
\vspace{2mm}
\caption{Gender Representation Metrics – MeToo Dataset}
\label{tab:metoo_metrics}
\end{table}

\section{Conclusions and Future Works}
Through this work, we propose a gender aware summarizer algorithm, \textit{EquiSumm} that can ensure representation of gender based opinions irrespective of the user identity. This is specifically relevant in current scenario when we prefer summarization algorithms to ensure fairness in summarization along with ageold objectives. We empirically evaluate the performance of the proposed model on two gender specific datasets to understand its effectiveness and compare with existing summarization approaches. Our preliminary results indicate high effectiveness of \textit{EquiSumm}. However, this is a preliminary work that we intend to extend such that it can capture and represent non‑binary or intersectional identities and include more real life datasets.
%Furthermore, we focus on exhaustive empirical analysis on several large scale datasets, that we have already collected and processed. %Since we focus exclusively on extractive summarizers rather than abstractive summarization, it is difficul t to comment on the effectiveness of the proposed model for abstractive summarization.

\clearpage

\bibliographystyle{splncs04}
\bibliography{sample-base}

\end{document}